\documentclass[utf8]{frontiersSCNS} % for Science, Engineering and Humanities and Social Sciences articles
\usepackage{url,lineno,microtype}
\usepackage{hyperref}
\usepackage{caption}
\usepackage{subcaption}
\usepackage[onehalfspacing]{setspace}
\usepackage{natbib}
\usepackage{graphicx}

\usepackage{multirow}
\usepackage{tabularx}

\def\firstAuthorLast{Roy {et~al.}} %use et al only if is more than 1 author
\def\Authors{Deboleena Roy\,$^{*}$, Priyadarshini Panda\, and Kaushik Roy}
\def\Address{Department of Electrical and Computer Engineering, Purdue University, West Lafayette, IN, USA}
\def\corrAuthor{Mailbox 312, EE Building, \\450 Northwestern Avenue, \\West Lafayette 47907, IN, USA}

\def\corrEmail{roy77@purdue.edu}

\begin{document}
\onecolumn
\firstpage{1}

\title[Synthesizing Images from Spatio-Temporal Representations]{Synthesizing Images from Spatio-Temporal Representations using Spike-based Backpropagation} 

\author[\firstAuthorLast ]{\Authors} %This field will be automatically populated
\address{} %This field will be automatically populated
\correspondance{} %This field will be automatically populated

\extraAuth{}% If there are more than 1 corresponding author, comment this line and uncomment the next one.
%\extraAuth{corresponding Author2 \\ Laboratory X2, Institute X2, Department X2, Organization X2, Street X2, City X2 , State XX2 (only USA, Canada and Australia), Zip Code2, X2 Country X2, email2@uni2.edu}

\maketitle

\begin{abstract}

%%% Leave the Abstract empty if your article does not require one, please see the Summary Table for full details.
\section{}
Spiking neural networks (SNNs) offer a promising alternative to current artificial neural networks to enable low-power event-driven neuromorphic hardware. Spike-based neuromorphic applications require processing and extracting meaningful information from spatio-temporal data, represented as series of spike trains over time. In this paper, we propose a method to synthesize images from multiple modalities in a spike-based environment. We use spiking auto-encoders to convert image and audio inputs into compact spatio-temporal representations that is then decoded for image synthesis. For this, we use a direct training algorithm that computes loss on the membrane potential of the output layer and back-propagates it by using a sigmoid approximation of the neuron's activation function to enable differentiability. The spiking autoencoders are benchmarked on MNIST and Fashion-MNIST and achieve very low reconstruction loss, comparable to ANNs. Then, spiking autoencoders are trained to learn meaningful spatio-temporal representations of the data, across the two modalities - audio and visual. We synthesize images from audio in a spike-based environment by first generating, and then utilizing such shared multi-modal spatio-temporal representations. Our audio to image synthesis model is tested on the task of converting TI-46 digits audio samples to MNIST images. We are able to synthesize images with high fidelity and the model achieves competitive performance against ANNs.
\end{abstract}

\section{Introduction}
In recent years, Artificial Neural Networks (ANNs) have become powerful computation tools for complex tasks such as pattern recognition, classification and function estimation problems \citep{lecun2015deep}. They have an ``activation" function in their compute unit, also know as a neuron. These functions are mostly \textit{sigmoid}, \textit{tanh}, or \textit{ReLU} \citep{nair2010rectified} and are very different from a biological neuron. Spiking neural networks (SNNs), on the other hand,  are recognized as the ``third generation of neural networks" \citep{maass1997networks}, with their ``spiking" neuron model much closely mimicking a biological neuron. They have a more biologically plausible architecture that can potentially achieve high computational power and efficient neural implementation \citep{ghosh2009spiking,maass2015spike}. 

%However, training methods for these spiking neural networks \citep{jin2018hybrid,sengupta2018going} are still in an early development stage, where each method comes with its own advantages and challenges.

For any neural network, the first step of learning is the ability to encode the input into meaningful representations. Autoencoders are a class of neural networks that can learn efficient data encodings in an unsupervised manner \citep{vincent2008extracting}. Their two-layer structure makes them easy to train as well. Also, multiple autoencoders can be trained separately and then stacked to enhance functionality \citep{masci2011stacked}. In the domain of SNNs as well, autoencoders provide an exciting opportunity for implementing unsupervised feature learning \citep{panda2016unsupervised}. Hence, we use autoencoders to investigate how input spike trains can be processed and encoded into meaningful hidden representations in a spatio-temporal format of output spike trains which can be used to recognize and regenerate the original input.

Generally, autoencoders are used to learn the hidden representations of data belonging to one modality only. However, the information surrounding us presents itself in multiple modalities - vision, audio, and touch. We learn to associate sounds, visuals and other sensory stimuli to one another. For example, an ``apple'' when shown as an image, or as text, or heard as an audio, holds the same meaning for us. A better learning system is one that is capable of learning shared representation of multimodal data \citep{srivastava2012learning}. \citet{wysoski2010evolving} proposed a bimodal SNN model that performs person authentication using speech and visual (face) signals. STDP-trained networks on bimodal data have exhibited better performance \citep{rathi2018stdp}. In this work, we explore the possibility of two sensory inputs - audio and visual, of the same object, learning a shared representation using multiple autoencoders, and then use this shared representation to synthesize images from audio samples.

To enable the above discussed functionalities, we must look at a way to train these spiking autoencoders. While several prior works exist in training these networks, each comes with its own advantages and drawbacks. One way to train spiking autoencoders is by using Spike Timing Dependent Plasticity (STDP) \citep{sjostrom2010spike}, an unsupervised local learning rule based on spike timings, such as \citet{burbank2015mirrored} and \citet{tavanaei2018representation}. However, STDP, being unsupervised and localized, still fails to train SNNs to perform at par with ANNs.  Another approach is derived from ANN backpropagation; the average firing rate of the output neurons is used to compute the global loss \citep{bohte2002error,lee2016training}. Rate-coded loss fails to include spatio-temporal information of the network, as the network response is accumulated over time to compute the loss. \citet{wu2018spatio} applied backpropagation through time (BPTT) \citep{werbos1990backpropagation}, while \citet{jin2018hybrid} proposed a hybrid backpropagation technique to incorporate the temporal effects. Very recently \cite{wu2018direct} demonstrated direct training of deep SNNs in a Pytorch based implementation framework. However, it continues to be a challenge to accurately map the time-dependent neuronal behavior with a time-averaged rate coded loss function.

In a network trained for classification, an output layer neuron competes with its neighbors for the highest firing rate, which translates into the class label, thus making rate-coded loss a requirement. However, the target for an autoencoder is very different. The output neurons are trained to regenerate the input neuron patterns. Hence, they provide us with an interesting opportunity where one can choose not to use rate-coded loss. Spiking neurons have an internal state, referred to as the membrane potential ($V_{mem}$), that regulates the firing rate of the neuron. The $V_{mem}$ changes over time depending on the input to the neuron, and whenever it exceeds a threshold, the neuron generates a spike. \citet{panda2016unsupervised} first presented a backpropagation algorithm for spiking autoencoders that uses $V_{mem}$ of the output neurons to compute the loss of the network. They proposed an approximate gradient descent based algorithm to learn hierarchical representations in stacked convolutional autoencoders. For training the autoencoders in this work, we compute the loss of the network using $V_{mem}$ of the output neurons, and we incorporate BPTT \citep{werbos1990backpropagation} by unrolling the network over time to compute the gradients. 

In this work, we demonstrate that in a spike-based environment, inputs can be transformed into compressed spatio-temporal spike maps, which can be then be utilized to reconstruct the input later, or can be transferred across network models, and data modalities. We train and test spiking autoencoders on MNIST and Fashion-MNIST dataset. We also present an audio-to-image synthesis framework, composed of multi-layered fully-connected spiking neural networks. A spiking autoencoder is used to generate compressed spatio-temporal spike maps of images (MNIST). A spiking audiocoder then learns to map audio samples to these compressed spike map representations, which are then converted back to images with high fidelity using the spiking autoencoder. To the best of our knowledge, this is the first work to perform audio to image synthesis in a spike-based environment.

The paper is organized in the following manner: In Section 2, the neuron model, the network structure and notations are introduced. The backpropagation algorithm is explained in detail. This is followed by Section 3 where the performance of these spiking autoencoders is evaluated on MNIST \citep{lecun1998gradient} and Fashion-MNIST \citep{xiao2017fashion} datasets. We then setup our Audio to Image synthesis model and evaluate it for converting TI-46 digits audio samples to MNIST images. Finally, in Section 4, we conclude the paper with discussion on this work and its future prospects.

\section{Learning Spatio-Temporal Representations using Spiking Autoencoders}
\label{VmemBackpropagationAlgorithm}

In this section, we understand the spiking dynamics of the autoencoder network and mathematically derive the proposed training algorithm, a membrane-potential based backpropagation. 

% add a an intro paragraph about importance of spatio-temporal representations

\subsection{Input Encoding and Neuron Model}

\begin{figure}[h]
\centering
\includegraphics[width=0.9\textwidth]{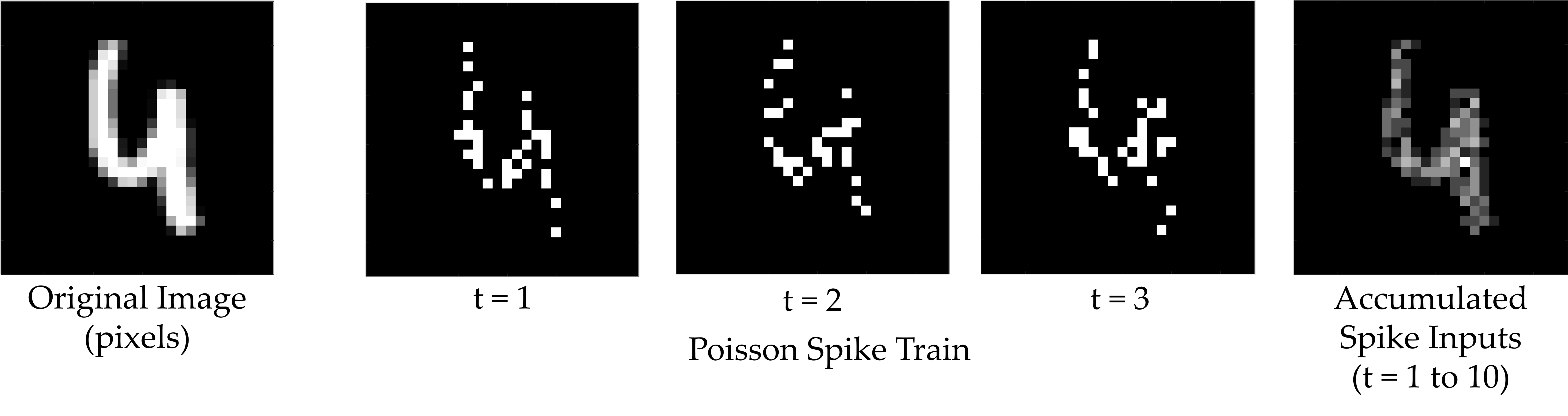}
\caption{The input image is converted into a spike map over time. At each time step neurons spike with a probability proportional to the corresponding pixel value at their location. These spike maps, when summed over several time steps, reconstruct the original input} \label{fig:input_to_spike}
\end{figure}

A spiking neural network differs from a conventional ANN in two main aspects - inputs and activation functions. For an image classification task, for example, an ANN would typically take the raw pixel values as input. However, in SNNs, inputs are binary spike events that happen over time. There are several methods for input encoding in SNNs currently in use, such as rate encoding, rank order coding and temporal coding \citep{wu2007learning}. One of the most common methods is rate encoding, where each pixel is mapped to a neuron that produces a Poisson spike train, and its firing rate is proportional to the pixel value. In this work, every pixel value of $0-255$ is scaled to a value between $[0,1]$ and a corresponding Poisson spike train of fixed duration, with a pre-set maximum firing rate, is generated (Fig.\ref{fig:input_to_spike}). 

The neuron model is that of a leaky integrate-and-fire (LIF) neuron. The membrane potential ($V_{mem}$) is the internal state of the neuron that gets updated at each time step based on the input of the neuron, $Z^{[t]}$ (eq. \ref{eq:vmem}).The output activation ($A^{[t]}$) of the neuron depends on whether  $V_{mem}$ reaches a threshold ($V_{th}$) or not. At any time instant, the output of the neuron is 0 unless the following condition is fulfilled, $V_{mem} \geq V_{th}$ (eq. \ref{eq:activation}). The leak factor is determined by a constant $\alpha$. After a neuron spikes, it's membrane potential is reset to 0. Fig. \ref{fig:ae_network_neuron_model}B illustrates a typical neuron's behavior over time.
\begin{equation}
\label{eq:vmem}
V_{mem}^{[t]} = (1-\alpha) V_{mem}^{[t-1]} + Z^{[t]}
\end{equation}
\begin{equation}
\label{eq:activation}
A^{[t]} = 
\begin{cases}
0, & V_{mem}^{[t]} < V_{th} \\
1, & V_{mem}^{[t]} \ge V_{th}
\end{cases}
\end{equation}

The activation function (eq. \ref{eq:activation}), which is a clip function, is non-differentiable with respect to $V_{mem}$, and hence we cannot take its derivative during backpropagation.  Several works use various approximate pseudo-derivatives, such as piece-wise linear \citep{esser2015backpropagation}, and exponential derivative \citep{shrestha2018slayer}. As mentioned in \citep{shrestha2018slayer}, the probability density function of the switching activity of the neuron with respect to its membrane potential can be used to approximate the clip function. It has been observed that biological neurons are noisy and exhibit a probabilistic switching behaviour \citep{nessler2013bayesian, benayoun2010avalanches}, which can be modeled as having a sigmoid-like characterstic \citep{sengupta2016probabilistic}. Thus, for backpropagation, we approximate the clip function (eq. \ref{eq:activation}) with a sigmoid which is centered around $V_{th}$, and thereby, the derivative of $A^{[t]}$ is approximated as the derivative of the sigmoid, ($A_{apx}^{[t]}$)  (eq. \ref{eq:act:approximate1}, \ref{eq:act:approximate2}). 
%Previously, \citet{sengupta2016probabilistic} used sigmoid activation function to train an ANN, and then performed ANN-SNN conversion, where the spiking neuron was a stochastic sigmoid. 
\begin{equation}
\label{eq:act:approximate1}
A_{apx}^{[t]} = \frac{1}{1+\exp(-(V_{mem}^{[t]}-V_{th}))}
\end{equation}
\begin{equation}
\label{eq:act:approximate2}
\frac{\partial A^{[t]}}{\partial V_{mem}^{[t]}} \approx \frac{\partial A^{[t]}_{apx}}{\partial V_{mem}^{[t]}} = \frac{\exp(-(V_{mem}^{[t]}-V_{th}))}{(1+\exp(-(V_{mem}^{[t]}-V_{th})))^2}
\end{equation}

\subsection{Network Model}

\begin{figure}[h]
\centering
\includegraphics[width=\linewidth]{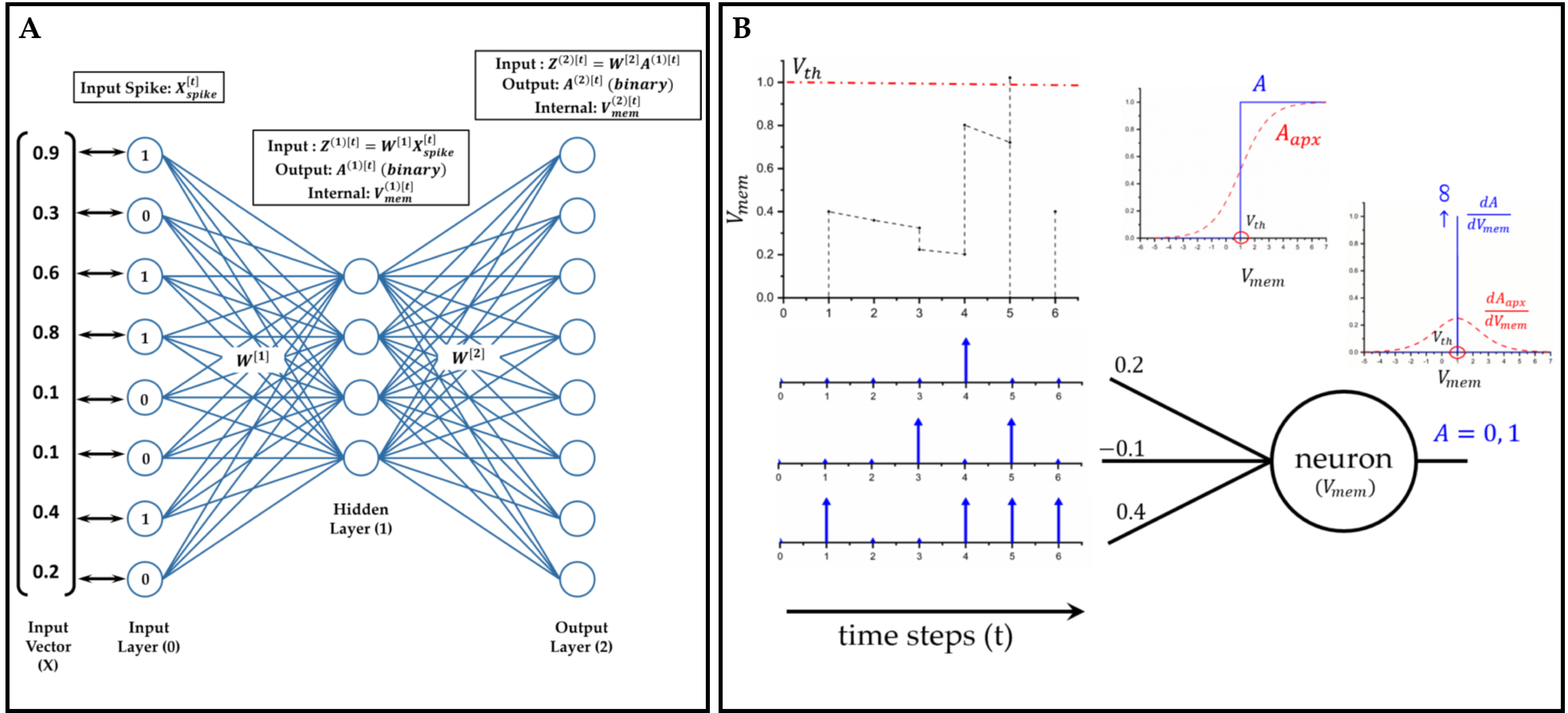}
\caption{The dynamics of a spiking neural network (SNN): \textbf{(A)} A two layer feed-forward SNN at any given arbitrary time instant. The input vector is mapped one-to-one to the input neurons ($layer^{(0)}$). The input value governs the firing rate of the neuron, i.e. number of times the neuron output is 1 in a given duration. \textbf{(B)} A leaky integrate and fire (LIF) neuron model with 3 synapses/weights at its input. The membrane potential of the neuron integrates over time (with leak). As soon as it crosses $V_{th}$, the neuron output changes to 1, and $V_{mem}$ is reset to 0. For taking derivative during backpropagation, a sigmoid approximation is used for the neuron activation}
\label{fig:ae_network_neuron_model}
\end{figure} 

We define the autoencoder as a two layer fully connected feed-forward network. To evaluate our proposed training algorithm, we have used two datasets - MNIST \citep{lecun1998gradient} and Fashion MNIST \citep{xiao2017fashion}. The two datasets have the same input size, a $28\times28$ gray-scale image. Hence, the input and the output layers of their networks have $784$ neurons each. The number of $layer^{(1)}$ neurons is different for the two datasets. The input neurons ($layer^{(0)}$) are mapped to the image pixels in a one-to-one manner and generate the Poisson spike trains. The autoencoder trained on MNIST later used as one of the building blocks of the audio-to-image synthesis network. The description of the network and the notation used throughout the paper is given in Fig. \ref{fig:ae_network_neuron_model}A.

\subsection{Backpropagation using Membrane Potential} 
\label{sec:backpropagation}
In this work, loss is computed using the membrane potential of output neurons at every time step and then it's gradient with respect to weights is backpropagated for weight update. The input image is provided to the network as 784$\times$1 binary vector over $T$ time steps, represented as $X_{spike}^{(t)}$. At each time step the desired membrane potential of the output layer is calculated (eq. \ref{eq:target_vmem}). The loss is the difference between the desired membrane potential and the actual membrane potential of the output neurons. Additionally a masking function is used that helps us focus on specific neurons at a time. The mask used here is bitwise \textit{XOR} between expected spikes ($X_{spike}^{[t]}$) and output spikes ($A^{(2)[t]}$) at a given time instant. The mask only preserves the error of those neurons that either were supposed to spike but did not spike, or were not supposed to spike, but spiked. It sets the loss to be zero for all other neurons. We observed that masking is essential for training in spiking autoencoder as shown in Fig. \ref{fig:ae_mnist_mask_timesteps}A

\begin{equation}
\label{eq:target_vmem}
O^{[t]} = V_{th}.^{*}X_{spike}^{[t]}
\end{equation}
\begin{equation}
mask = bitXOR(X_{spike}^{[t]}, A^{(2)[t]})
\end{equation}
\begin{equation}
\label{eq:error}
Error = E = mask.^{*}(O^{[t]} - V_{mem}^{(2)[t]})
\end{equation}
\begin{equation}
Loss = L = \frac{1}{2}|E|^{2}
\end{equation}
The weight gradients, $\frac{\partial L}{\partial W}$, are computed by back-propagating loss in the two layer network as depicted in Fig. \ref{fig:ae_network_neuron_model}A. We derive the weight gradients below. 
\begin{equation}
\label{eq:dl/dvmem-2}
\frac{\partial L}{\partial V_{mem}^{(2)[t]}} = -E 
\end{equation}
From eq. \ref{eq:vmem},
\begin{equation}
\label{eq:dvmem-2/dw-2}
\frac{\partial V_{mem}^{(2)[t]}}{\partial W^{(2)}} = (1-\alpha) \frac{\partial V_{mem}^{(2)[t-1]}}{\partial W^{(2)}} + \big[A^{(1)[t]}\big]^{T} .
\end{equation}
The derivative is dependent not only on the current input ($A^{(1)[t]}$), but also on the state from previous time step ($V_{mem}^{(2)[t-1]}$). 
\newline
Next we apply chain rule on eq. \ref{eq:dl/dvmem-2} - \ref{eq:dvmem-2/dw-2},
\begin{equation}
\label{eq:dl/dw2}
\frac{\partial L}{\partial W^{(2)}} = \frac{\partial L}{\partial V_{mem}^{(2)[t]}} \frac{\partial V_{mem}^{(2)[t]}}{\partial W^{(2)}}= -E\bigg[(1 - \alpha) \frac{\partial V_{mem}^{(2)[t-1]}}{\partial W^{(2)}} + \big[A^{(1)[t]}\big]^{T}\bigg] ,
\end{equation} 
from eq. \ref{eq:vmem}, 
\begin{equation}
\label{eq:dvmem-2/dz-2}
\frac{\partial V_{mem}^{(2)[t]}}{\partial Z^{(2)[t]}} = I  ,
\end{equation}
from \ref{eq:dl/dvmem-2} and \ref{eq:dvmem-2/dz-2}, we obtain the local error of $layer^{(2)}$ with respect to the overall loss which is backpropagated to $layer^{(1)}$,
\begin{equation}
\label{eq:delta_2}
\delta_{2} = \frac{\partial L}{\partial Z^{(2)[t]}} = I(-E) = -E ,
\end{equation}
next, the gradients for $layer^{(1)}$ are calculated,
\begin{equation}
\label{eq:dz-2/da-1}
\frac{\partial Z^{(2)[t]}}{\partial A^{(1)[t]}} = W^{(2)} ,
\end{equation}
from eq. \ref{eq:act:approximate1} - \ref{eq:act:approximate2},
\begin{equation}
\label{eq:da-1/dvmem-1}
\frac{\partial A^{(1)[t]}}{\partial V_{mem}^{(1)[t]}} \approx \frac{\partial A^{(1)[t]}_{apx}}{\partial V_{mem}^{(1)[t]}} = \frac{\exp(-(V_{mem}^{(1)[t]}-V_{th}))}{(1+\exp(-(V_{mem}^{(1)[t]}-V_{th})))^2} ,
\end{equation}
from eq. \ref{eq:vmem},
\begin{equation}
\label{eq:dvmem-1/dw-1}
\frac{\partial V_{mem}^{(1)[t]}}{\partial W^{(1)}} = (1-\alpha) \frac{\partial V_{mem}^{(1)[t-1]}}{\partial W^{(1)}} + \big[X_{spike}^{[t]}\big]^{T} ,
\end{equation}
from \ref{eq:delta_2} - \ref{eq:dvmem-1/dw-1}, 
\begin{equation}
\label{eq:dl/dw1}
\frac{\partial L}{\partial W^{(1)}} = \frac{\partial L}{\partial V_{mem}^{(1)[t]}} \frac{\partial V_{mem}^{(1)[t]}}{\partial W^{(1)}} = \bigg[\big[W^{(2)}\big]^{T}\delta_{2} \circ  \frac{\partial A^{(1)[t]}}{\partial V_{mem}^{(1)[t]}}\bigg]\bigg[(1-\alpha) \frac{\partial V_{mem}^{(1)[t-1]}}{\partial W^{(1)}} + \big[X_{spike}^{[t]}\big]^{T}\bigg] .
\end{equation}
Thus, equations \ref{eq:dl/dw2} and \ref{eq:dl/dw1} show how gradients of the loss function with respect to weights are calculated. For weight update, we use mini-batch gradient descent and a weight decay value of 1e-5. We implement Adam optimization \citep{kingma2014adam}, but the first and second moments of the weight gradients are averaged over time steps per batch (and not averaged over batches). We store $\frac{\partial V_{mem}^{(l)[t]}}{\partial W^{(l)}}$ of the current time step for use in next time step. The initial condition is, $\frac{\partial V_{mem}^{(l)[0]}}{\partial W^{(l)}} = 0$. If a neuron spikes, it's membrane potential is reset and therefore we reset $\frac{\partial V_{mem}^{(l,m)[t]}}{\partial W^{(l)}}$ to 0 as well, where $l$ is the layer number and $m$ is the neuron number. 

\section{Experiments}

\subsection{Regenerative Learning with Spiking Autoencoders}
\label{subsection:spiking_autoencoders}

\begin{figure}[h]
\centering
\includegraphics[width=\textwidth]{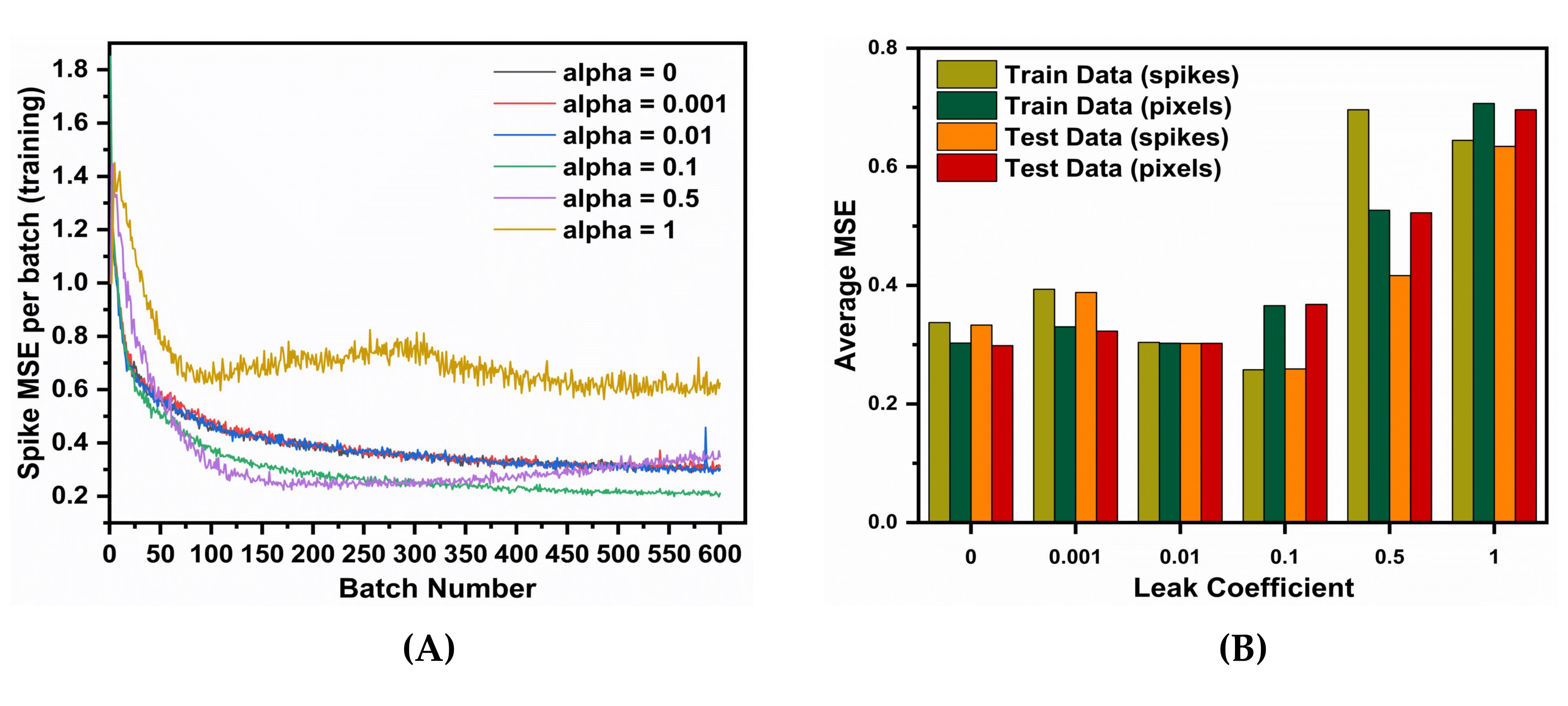}
\caption{The AE-SNN (784-196-784) is trained over MNIST (60,000 training samples, batch size = 100) for different leak coefficients ($\alpha$). \textbf{(A)} spike-based MSE (Mean Square Error) Reconstruction Loss per batch during training. \textbf{(B)} Average MSE over entire dataset after training}
\label{fig:ae_mnist_leak}
\end{figure}

\begin{figure}[h]
\centering
\includegraphics[width=\textwidth]{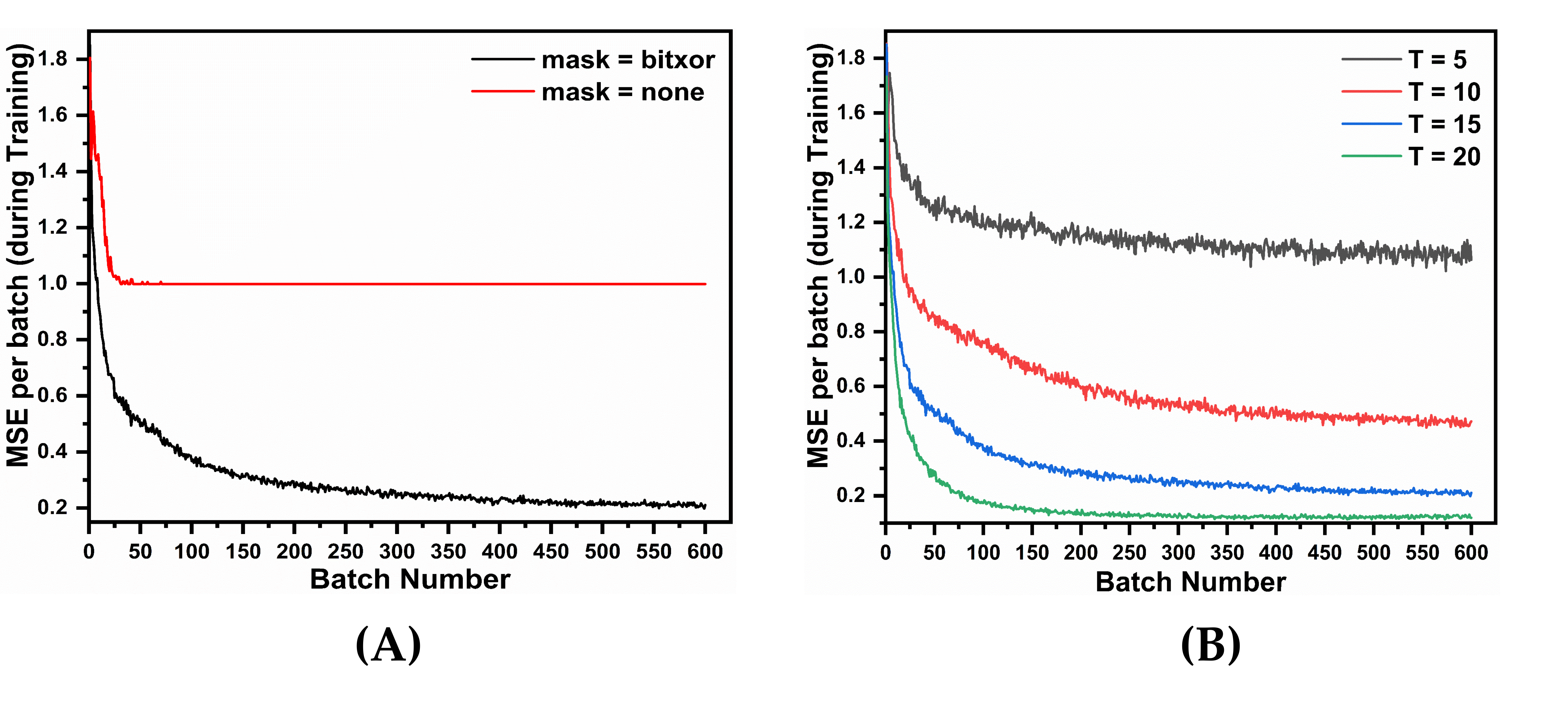}
\caption{The AE-SNN (784-196-784) is trained over MNIST (60,000 training samples, batch size = 100) and we study the impact of \textbf{(A)} mask, and \textbf{(B)} input spike train duration on the Mean Square Error (MSE) Reconstruction Loss}\label{fig:ae_mnist_mask_timesteps}
\end{figure}

\begin{figure}[h]
\centering
\includegraphics[width=\textwidth]{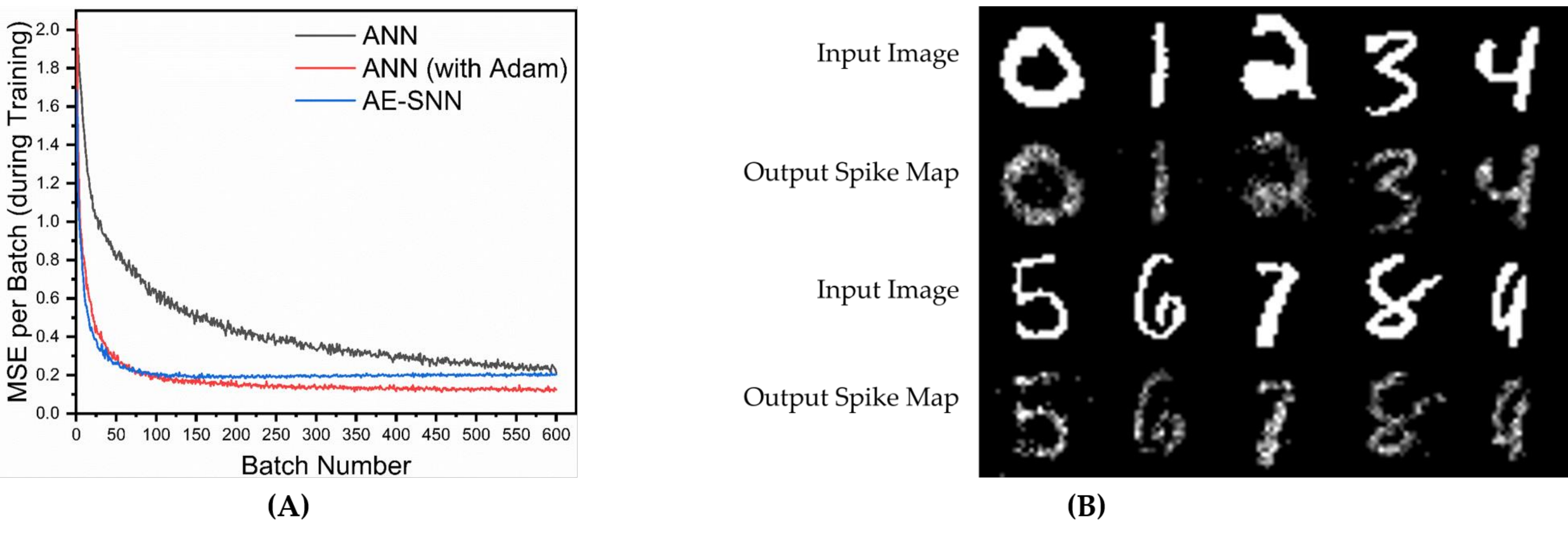}
\caption{AE-SNN trained on MNIST (training examples = 60,000, batch size = 100). \textbf{(A)} Spiking autoencoder (AE-SNN) versus AE-ANNs (trained with/without Adam). \textbf{(B)} Regenerated images from test set for AE-SNN (input spike duration = 15, leak = 0.1)}  
\label{fig:ae_mnist_ann_collage}
\end{figure}

\begin{figure}[h]
\centering
\includegraphics[width=\textwidth]{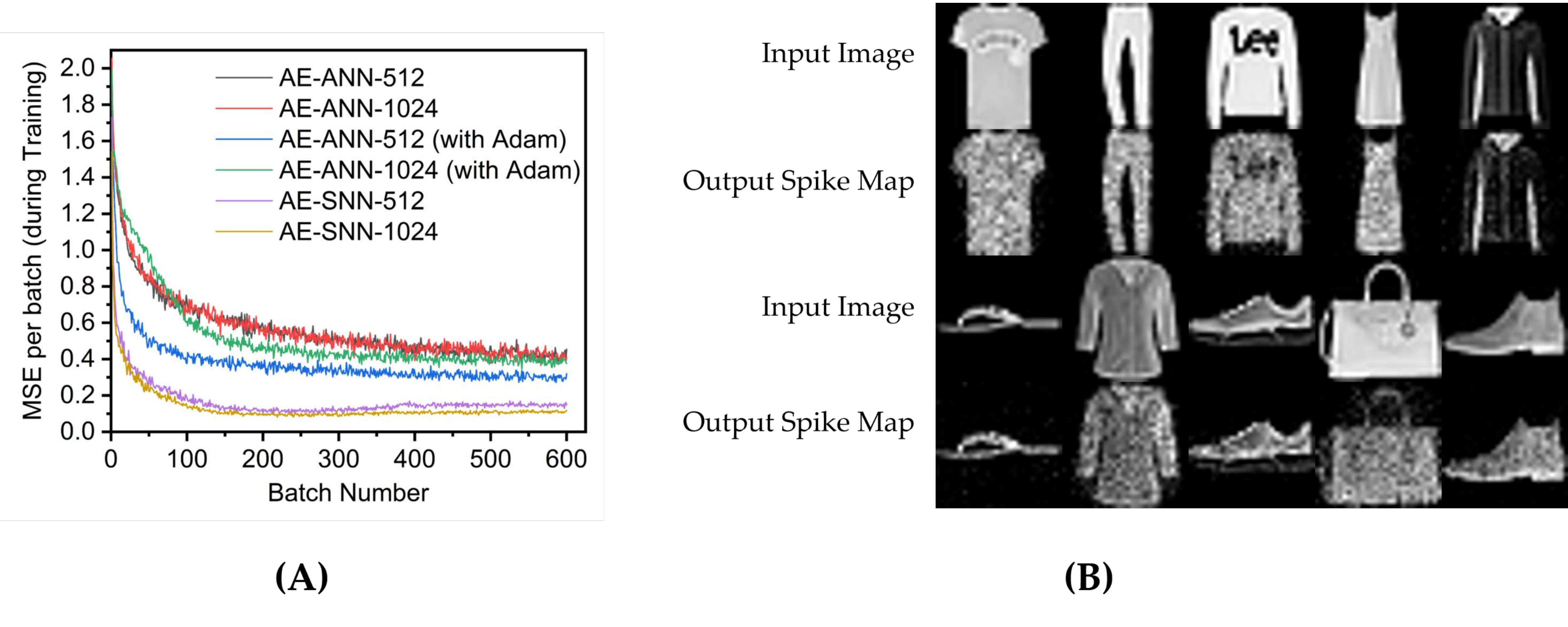}
\caption{AE-SNN trained on Fashion-MNIST (training examples = 60,000, batch size = 100) \textbf{(A)} AE-SNN (784$\times$(512/1024)$\times$784) versus AE-ANNs (trained with/without Adam, lr = 5e-3) \textbf{(B)} Regenerated images from test set for AE-SNN-1024}
\label{fig:ae_fmnist_ann_collage}
\end{figure}

\begin{figure}[h]
\centering
\includegraphics[width=\textwidth]{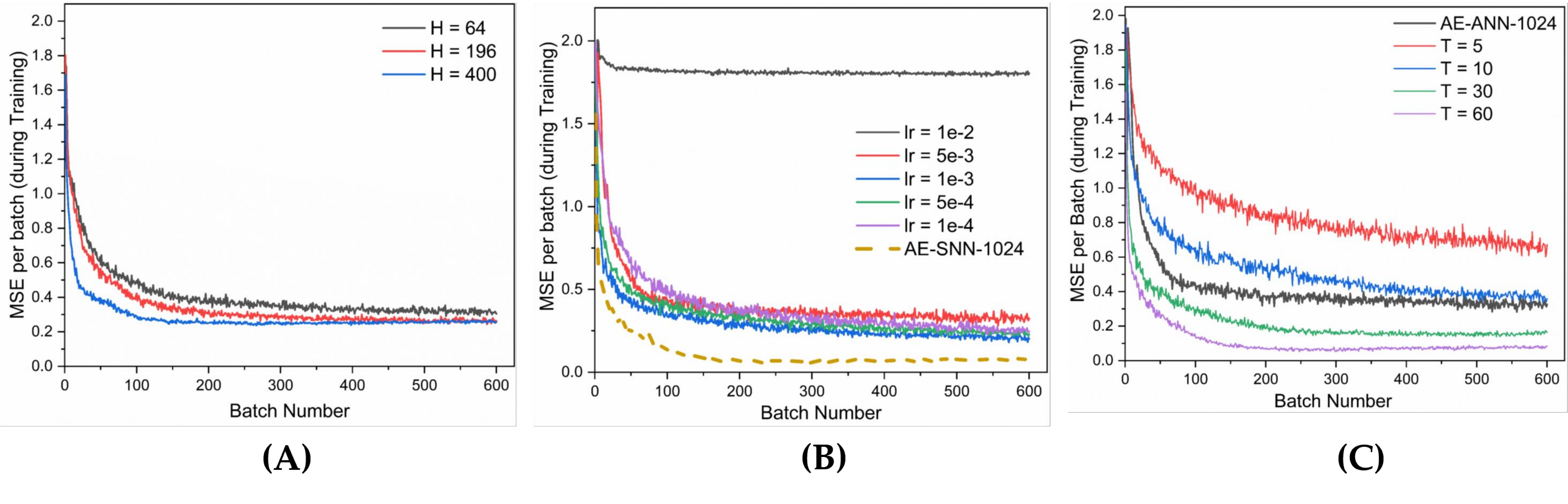}
\caption{\textbf{(A)} AE-SNN (784$\times$H$\times$784) trained on MNIST (training examples = 60,000, batch size = 100) for different hidden layer sizes = 64, 196, 400 \textbf{(B)} AE-ANN (784$\times$1024$\times$784) trained on Fashion-MNIST (training examples = 60,000, batch size = 100) with Adam optimization for various learning rates (lr). Baseline: AE-SNN trained with input spike train duration of 60 time steps. \textbf{(C)} AE-SNN (784$\times$1024$\times$784) trained on Fashion-MNIST (training examples = 60,000, batch size = 100) for varying input time steps, T = 15, 30, 60. Baseline: AE-ANN trained using Adam with lr = 5e-3}
\label{fig:rebuttal}
\end{figure}

For MNIST, a 784-196-784 fully connected network is used. The spiking autoencoder (AE-SNN) is trained for 1 epoch with a batch size of 100, learning rate 5e-4, and a weight decay of 1e-4.  The threshold ($V_{th}$) is set to 1. We define two metrics for network performance, Spike-MSE and MSE. Spike-MSE is the mean square error between the input spike map and the output spike map, both summed over the entire duration. MSE is the mean square error between the input image and output spike map summed over the entire duration. Both, input image and output map, are normalized, zero mean and unit variance, and then the mean square error is computed. The duration of inference is kept the same as the training duration of the network. 

It is observed in Fig. \ref{fig:ae_mnist_leak} that the leak coefficient plays an important role in the performance of the network. While a small leak coefficient improves performance, too high of a leak degrades it greatly. We use Spike-MSE as the comparison metric during training in Fig. \ref{fig:ae_mnist_leak}A, to observe how well the autoencoder can recreate the input spike train. In Fig. \ref{fig:ae_mnist_leak}B, we report two different MSEs, one computed against input spike map (spikes) and the other compared firing rate to pixel values (pixels), after normalizing both. For 'IF' neuron ($\alpha = 0$), the train data performs worse than test data, implying underfitting. At $\alpha$ set to $0.01$ we find the network having comparable performance between test and train datasets, indicating a good fit. At $\alpha = 0.1$, the Spike-MSE is lowest for both test and train data, however the MSE is higher. While the network is able to faithfully reconstruct the input spike pattern, the difference between Spike-MSE and regular MSE is because of the difference in actual pixel intensity and the converted spike maps generated by the poisson generator at the input. On further increasing the leak, there is an overall performance degradation on both test and train data. Thus, we observe that leak coefficient needs to be fine-tuned for optimal performance. Going forth, we set the leak coefficient at 0.1 for all subsequent simulations, as it gave the lowest train and test data MSE on direct comparison with input spike maps. 

Fig. \ref{fig:ae_mnist_mask_timesteps}A shows that using a mask function is essential for training this type of network. Without a masking function the training loss does not converge. This is because all of the 784 output neurons are being forced to have membrane potential of 0 or $V_{th}$, resulting in a highly constrained optimization space, and the network eventually fails to learn any meaningful representations. In the absence of any masking function, the sparsity of the error vector $E$ was less than 5\%, whereas, with the mask, the average sparsity was close to 85\%. This allows the optimizer to train the critical neurons and synapses of the network. The weight update mechanism learns to focus on correcting the neurons that do not fire correctly, which effectively reduces the number of learning variables, and results in better optimization. 

Another interesting observation was that increasing the duration of the input spike train improves the performance as shown in Fig.\ref{fig:ae_mnist_mask_timesteps}B. However, it comes at the cost of increased training time as backpropagation is done at each time step, as well as increased inference time.  We settle for an input time duration of 15 as a trade-off between MSE and time taken to train and infer for the next set of simulations. 

We also study the impact of hidden layer size for the reconstruction properties of the autoencoder. As shown in Fig. \ref{fig:rebuttal}A, as we increase the size of the network, the performance improves. However, this comes at the cost of increased network size, longer training time and slower inference. While one gets a good improvement when increasing hidden layer size from 64 to 196, the benefit diminishes as we increase the hidden layer size to 400 neurons. Thus for our comparison with ANNs, we use the 784$\times$196$\times$784 network. 

For comparison with ANNs, a network (AE-ANN) of same size (784$\times$196$\times$784) is trained with SGD, both with and without Adam optimizer \citep{kingma2014adam} on MNIST for 1 epoch with a learning rate of 0.1, batch size of 100, and weight decay of 1e-4. When training the AE-SNN, the first and second moments of the gradients are computed over sequential time steps within a batch (and not across batches). Thus it is not analogous to the AE-ANN trained with Adam, where the moments are computed over batches. Hence, we compare our network with both variants of the AE-ANNs, trained with and without Adam. The AE-SNN achieves better performance than the AE-ANN trained without Adam; however it lags behind the AE-ANN optimized with Adam as shown in Fig. \ref{fig:ae_mnist_ann_collage}A. Some of the reconstructed MNIST images are depicted in Fig. \ref{fig:ae_mnist_ann_collage}B. One important thing to note is that the AE-SNN is trained at every time step, hence there are 15$\times$ more backpropagation steps as compared to an AE-ANN. However at every backpropagation step, the AE-SNN only backpropagates the error vector of a single spike map, which is very sparse, and carries less information than the error vector of the AE-ANN. 

Next, the spiking autoencoder is evaluated on the Fashion-MNIST dataset \citep{xiao2017fashion}. It is similar to MNIST, and comprises of 28$\times$28 gray-scale images (60,000 training, 10,000 testing) of clothing items belonging to 10 distinct classes. We test our algorithm on two network sizes: 784-512-784 (AE-SNN-512) and 784-1024-784 (AE-SNN-1024). The AE-SNNs are compared against AE-ANNs of the same sizes (AE-ANN-512, AE-ANN-1024) in Fig. \ref{fig:ae_fmnist_ann_collage}A.  For the AE-SNNs, the duration of input spike train is 60, leak coefficient is 0.1, and learning rate is set at 5e-4. The networks are trained for 1 epoch, with a batch size of 100. The longer the spike duration, the better would be the spike image resolution. For a duration of 60 time steps, a neuron can spike anywhere between zero to 60 times, thus allowing 61 gray-scale levels. Some of the generated images by AE-SNN-1024 are displayed in Fig. \ref{fig:ae_fmnist_ann_collage}B. The AE-ANNs are trained for 1 epoch, batch size 100, learning rate 5e-3 and weight decay 1e-4. 

For Fashion-MNIST, the AE-SNNs exhibited better performance than AE-ANNs as shown in Fig. \ref{fig:ae_fmnist_ann_collage}A. We varied the learning rate for AE-ANN, and the AE-SNN still outperformed it's ANN counterpart (Fig. \ref{fig:rebuttal}B). This is an interesting observation, where the better performance comes at the increased effort of per-batch training. Also it exhibits such behavior on only this dataset, and not on MNIST (Fig.\ref{fig:ae_mnist_ann_collage}A). The spatio-temporal nature of training over each time step could possibly train the network to learn the details in an image better. Spiking Neural Networks have an inherent sparsity in them which could possibly acts like a dropout regularizer \citep{srivastava2014dropout}. Also, in case of AE-SNN, the update is made at every time step (60 updates per batch), in contrast to ANN where there is one update for one batch. We evaluated AE-SNN for shorter time steps, and observe that for smaller time steps (T = 5, 10), AE-SNN performs worse than AE-ANN (Fig. \ref{fig:rebuttal}C). The impact of time steps is greater for Fashion-MNIST, as compared to MNIST (Fig. \ref{fig:ae_mnist_mask_timesteps}B), as Fashion-MNIST data has more grayscale levels than the near-binary MNIST data. We also observed that, for both datasets, MNIST and Fashion-MNIST, the AE-SNN converges faster than AE-ANNs trained without Adam, and converges at almost the same time as an AE-ANN trained with Adam. The proposed spike-based backpropagation algorithm is able to bring the AE-SNN performance at par, and at times even better, than AE-ANNs.

\subsection{Audio to Image Synthesis using Spiking Auto-Encoders}

\subsubsection{Dataset} 
\label{subsubsection:audio_to_image_dataset}
For the audio to image conversion task, we use two standard datasets, the 0-9 digits subset of TI-46 speech corpus \citep{liberman1993ti} for audio samples, and MNIST dataset \citep{lecun1998gradient} for images. The audio dataset has read utterances of 16 speakers for the 10 digits, with a total 4136 audio samples. We divide the audio samples into 3500 train samples and 636 test samples, maintaining an 85\%/15\% train/test ratio. For training, we pair each audio sample with an image. We chose two ways of preparing these pairs, as described below:  
\begin{enumerate}
\item \textbf{Dataset A}: $10$ unique images of the $10$ digits is manually selected ($1$ image per class) and audio samples are paired with the image belonging to their respective classes (one-image-per-audio-class). All audio samples of a class are paired with the identical image of a digit belonging to that class. 
\item \textbf{Dataset B}: Each audio sample of the training set is paired with a randomly selected image (of the same label) from the MNIST dataset (one-image-per-audio-sample). Every audio sample is paired with a unique image of the same class. 
\end{enumerate}

The testing set is same for both Dataset A and B, comprising of 636 audio samples. All the audio clips were preprocessed using Auditory Toolbox \citep{slaney1998auditory}. They were converted to spectrograms having 39 frequency channels over 1500 time steps. The spectrogram is then converted into a 58500$\times$1 vector of length 58500. This vector is then mapped to the input neurons ($layer^{(0)}$) of the audiocoder, which then generate Poisson spike trains over the given training interval.
\subsubsection{Network Model}

\begin{figure}[h]
\centering
\includegraphics[width=0.9\linewidth]{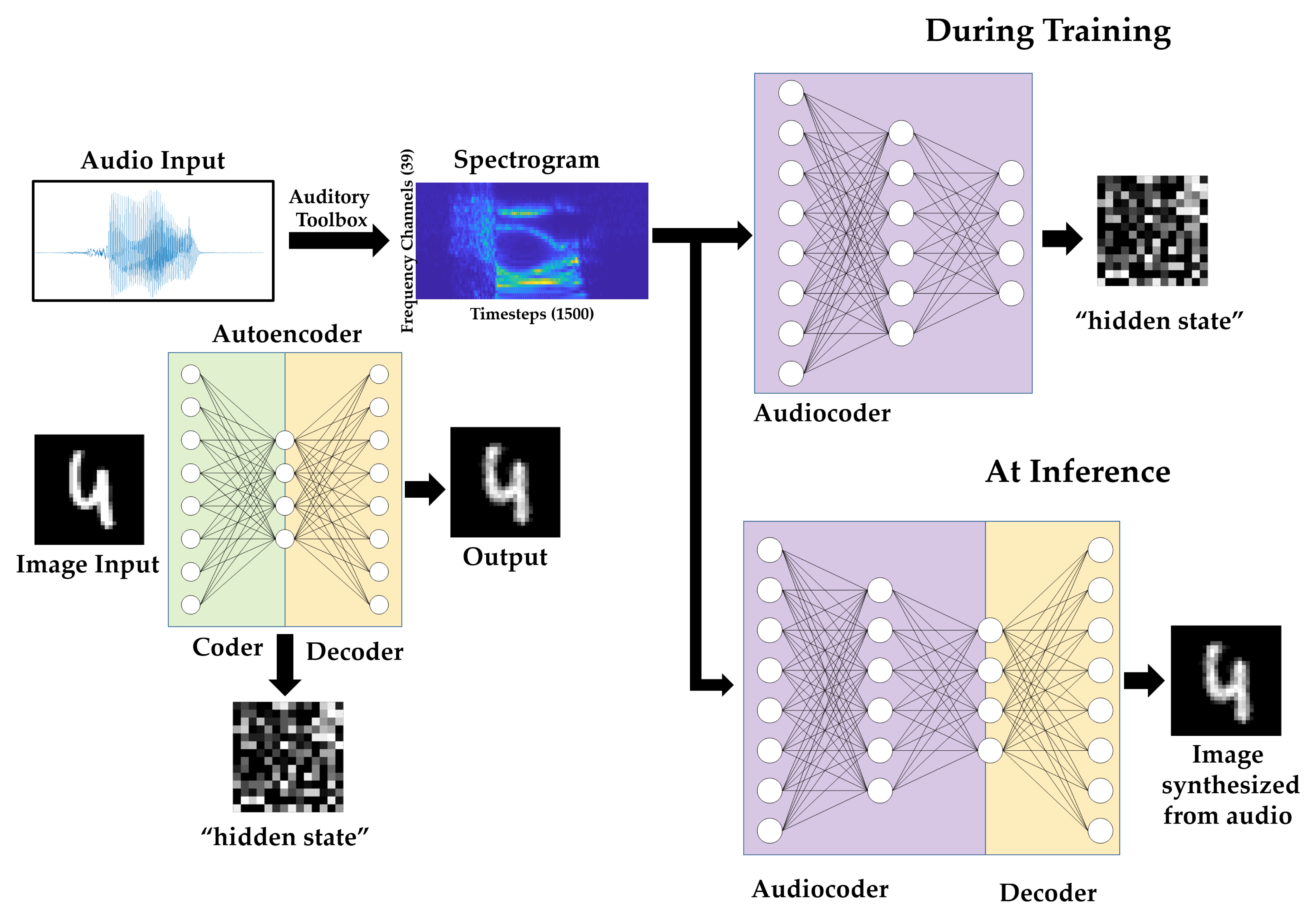}
\caption{Audio to Image synthesis model using an Autoencoder trained on MNIST images, and an Audiocoder trained to convert TI-46 digits audio samples into corresponding hidden state of the MNIST images.} \label{fig:audio_to_image_synthesis_model}
\end{figure} 

The principle of stacked autoencoders is used to perform audio-to-image synthesis. An autoencoder is built of two sets of weights; the $layer^{(1)}$ weights ($W^{(1)}$) encodes the information into a ``hidden state'' of a different dimension, and the second layer ($W^{(2)}$) decodes it back to it's original representation. We first train a spiking autoencoder on MNIST dataset. We use the AE-SNN as trained in Fig. \ref{fig:ae_mnist_ann_collage}A. Using $layer^{(1)}$ weights ($W^{[1]}$) of this AE-SNN, we generate ``hidden-state'' representations of the images belonging to the training set of the multimodal dataset. These hidden-state representations are spike trains of a fixed duration. Then we construct an audiocoder: a two layer spiking network that converts spectrograms to this hidden state representation. The audiocoder is trained with membrane potential based backpropagation as described in Section \ref{sec:backpropagation}. The generated representation, when fed to the ``decoder'' part of the autoencoder, gives us the corresponding image. The network model is illustrated in Fig. \ref{fig:audio_to_image_synthesis_model}

\subsubsection{Results}

The MNIST autoencoder (AE-SNN) used for audio-to-image synthesis task is trained using the following parameters: batch size of 100, learning rate 5e-4, leak coefficient 0.1, weight decay 1e-4, input spike train duration 15, and number of epochs 1, as used in section 3.1. We use Dataset A and Dataset B (as described in section \ref{subsubsection:audio_to_image_dataset}) to train and evaluate our audio-to-image synthesis model. The images that were paired with the training audio samples are converted to Poisson spike trains (duration 15 time steps) and fed to the AE-SNN, which generates a 196$\times$15 corresponding bitmap as the output of $layer^{(1)}$ (Fig. \ref{fig:ae_network_neuron_model}A). This spatio temporal representation is then stored. Instead of storing the entire duration of 15 time steps, one can choose to store a subset, such as first 5 or 10 time steps. We use $T_{h}$ to denote the saved hidden state's duration. 

This stored spike map serves as the target spike map for training the audiocoder (AC-SNN), which is a 58500$\times$2048$\times$196 fully connected network. The spectrogram (39$\times$1500) of each audio sample was converted to 58500$\times$1 vector which is mapped one-to-one to the input neurons($layer^{(0)}$). These input neurons then generate Poisson spike trains for 60 time steps.  The target map, of $T_{h}$ time steps, was shown repeatedly over this duration. The audiocoder (AC-SNN) is trained over 20 epochs, with a learning rate of 5e-5 and a leak coefficient of 0.1. Weight decay is set at 1e-4 and the batch size is 50. Once trained, the audiocoder is then merged with $W^{(2)}$ of AE-SNN to create the audio-to-image synthesis model (Fig. \ref{fig:audio_to_image_synthesis_model}). 

\begin{figure}[h!]
\centering
\includegraphics[width=\textwidth]{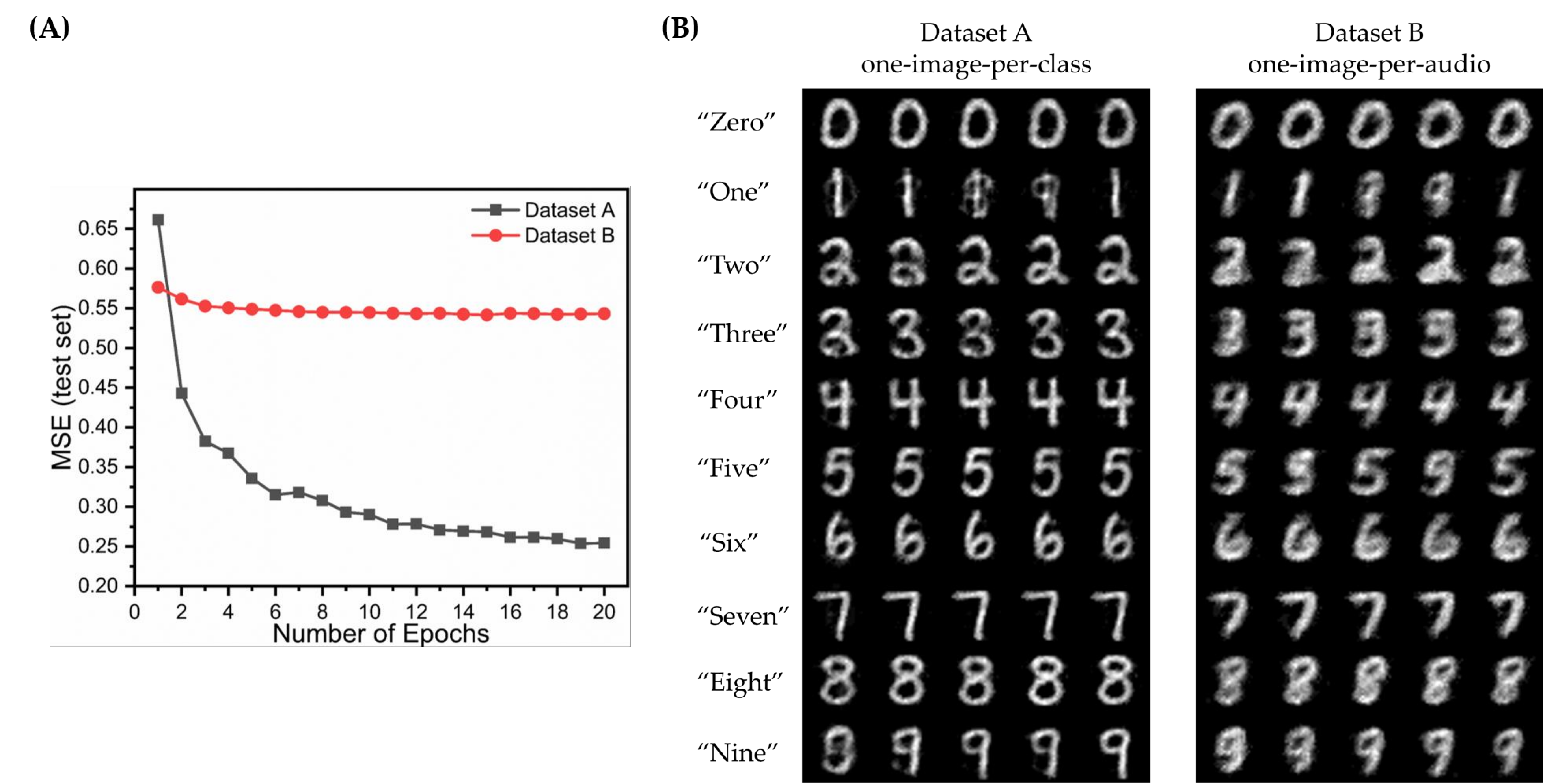}
\caption{The performance of the Audio to Image synthesis model on the two datasets - A and B ($T_{h}$ = 10)) \textbf{(A)} Mean square error loss (test set) \textbf{(B)} Images synthesized from different test audio samples (5 per class) for the two datasets A, and B}
\label{fig:ac_mnist_dataset_collage}
\end{figure}

For Dataset A, we compare the images generated by audio samples of a class against the MNIST image of that class to compute the MSE. In case of Dataset B, each audio sample of the train set is paired with an unique image. For calculating training set MSE, we compare the paired image and the generated image. For testing set, the generated image of an audio sample is compared with all the training images having the same label in the dataset, and the lowest MSE is recorded. The output spike map is normalized and compared with the normalized MNIST images, as was done previously. Our model gives lower MSE for Dataset A compared to Dataset B (Fig \ref{fig:ac_mnist_dataset_collage}A), as it is easier to learn just one representative image for a class, than unique images for every audio sample. The network trained with Dataset A generates very good identical images for audio samples belonging to a class. In comparison the network trained on Dataset B generates a blurry image, thus indicating that it has learned to associate the underlying shape and structure of the digits, but has not been able to learn finer details better. This is because the network is trained over multiple different images of the same class, and it learns what is common among them all. Fig. \ref{fig:ac_mnist_dataset_collage}B displays the generated output spike map for the two models trained over Dataset A and B for 50 different test audio samples (5 of each class). 

\begin{figure}[h!]
\centering
\includegraphics[width=\textwidth]{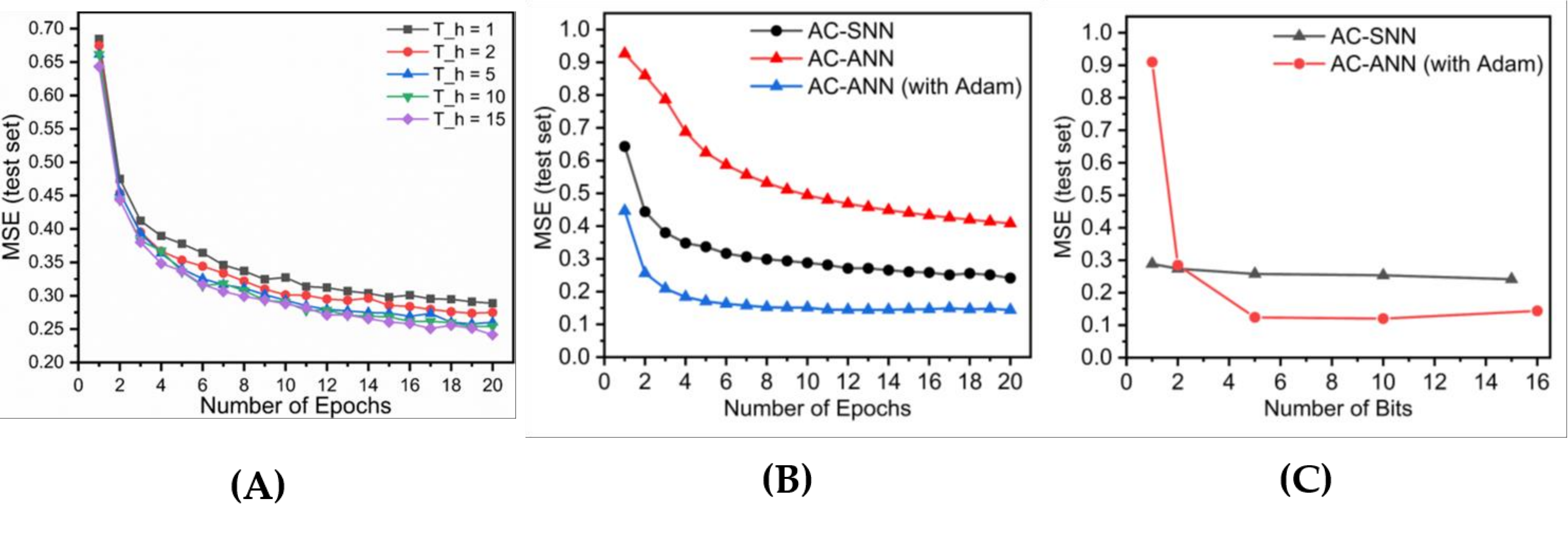}
\caption{The audiocoder (AC-SNN/AC-ANN) is trained over Dataset A, while the autoencoder (AE-SNN/AE-ANN) is fixed. MSE is reported on the overall audio-to-image synthesis model composed of AC-SNN/ANN and AE-SNN/ANN. \textbf{(A)} Reconstruction loss of the audio-to-image synthesis model for varying $T_{h}$ \textbf{(B)} Audiocoder performance AC-SNN ($T_{h} = 15$) vs AC-ANN (16 bit full precision) \textbf{(C)} Effect of training with reduced hidden state representation on AC-SNN and AC-ANN models}
\label{fig:ac_mnist_hs_ann_quant}    
\end{figure}

The duration ($T_{h}$) of stored ``hidden state'' spike train was varied from 15 to 10, 5, 2, and 1.  A spike map at a single time step is a 1-bit representation. The AE-SNN compresses an 784$\times$8 bit representation into 196$\times T_{h}$-bit representation. For $T_{h}$ = 15, 10, 5, 2, and 1, the compression is 2.1$\times$, 3.2$\times$, 6.4$\times$, 16$\times$ and 32$\times$ respectively. In Fig. \ref{fig:ac_mnist_hs_ann_quant}A we observe the reconstruction loss (test set) over epochs for training using different lengths of hidden state.  Even when the AC-SNN is trained with a much smaller ``hidden state'', the AE-SNN is able to reconstruct the images without much loss.  

For comparison, we initialize an ANN audiocoder (AC-ANN) of size 58500$\times$2048$\times$196. The AE-ANN trained over MNIST in section \ref{subsection:spiking_autoencoders} is used to convert the images of the multimodal dataset (A/B) to 196$\times$1 ``hidden state'' vectors. Each element of this vector is 16 bit full precision number. In case of AE-SNN, the ``hidden state'' is represented as a 196$\times T_{h}$ bit map. For comparison, we quantize the equivalent hidden state vector into $2^{T_{h}}$ levels. The AC-ANN is trained using these quantized hidden state representations with the following learning parameters: learning rate 1e-4, weight decay 1e-4, batch size 50, epochs 20. Once trained, the ANN audio-to-image synthesis model is built by combining AC-ANN and $layer^{(2)}$ weights ($W^{(2)}$) of AE-ANN. The AC-ANN is trained with/without Adam optimizer, and is paired with the AE-ANN trained with/without Adam optimizer respectively. In Fig. \ref{fig:ac_mnist_hs_ann_quant}B, we see that our spiking model achieves a performance in between the two ANN models, a trend we have observed earlier while training autoencoders on MNIST. In this case, the AC-SNN is trained with $T_{h}$ as 15, while AC-ANNs are trained without any output quantization; both are trained on Dataset A. In Fig. \ref{fig:ac_mnist_hs_ann_quant}C, we observe the impact of quantization for the ANN model and the corresponding impact of lower $T_{h}$ for SNN. For higher hidden state bit precision, the ANN model outperforms the SNN one. However for extreme quantization case, number of bits = 2, and 1, the SNN performs better. This could possibly be attributed to the temporal nature of SNN, where the computation is event-driven and spread out over several time steps. 

Note, all simulations were performed using MATLAB, which is a high level simulation environment. The algorithm, however, is agnostic of implementation environment from a functional point of view and can be easily ported to more traditional ML frameworks such as PyTorch or TensorFlow. 
\begin{table}[h]
\centering
\caption{Summary of results obtained for the 3 tasks - Autoencoder on MNIST, Autoencoder on Fashion-MNIST, and Audio to Image conversion (T = input duration for SNN)}
\centering
\begin{tabular}{|c|c|c|c|c|c|c|}
\hline
\multirow{2}{*}{Dataset} & \multirow{2}{*}{Network Size} & \multirow{2}{*}{Epochs} & \multirow{2}{*}{T} & \multicolumn{3}{c|}{Loss (MSE) (test)} \\ 
\cline{5-7} & & & & SNN & ANN & ANN (with Adam)\\ \hline
MNIST & 784-196-784 & 1 & 15 & 0.357 & 0.226 & \textbf{0.122}\\ \hline
\multirow{2}{*}{Fashion-MNIST} & 784-512-784 & 1 & 60 & \textbf{0.178} & 0.416 & 0.300 \\ 
\cline{2-7} & 784-1024-784 & 1 & 60 & \textbf{0.140} & 0.418 & 0.387 \\ \hline
Audio-to-Image A  & 58500-2048-196/196-784 & 20 & 30 & 0.254 & 0.408 & \textbf{0.144} \\ \hline
Audio-to-Image B & 58500-2048-196/196-784 & 20 & 30 & \textbf{0.543} & 0.611 & 0.556 \\ \hline
\end{tabular}
\label{table:summary}
\end{table}

\section{Discussion and Conclusion}
In this work, we propose a method to synthesize images in spike-based environment. In Table \ref{table:summary}, we have summarized the results of training autoencoders and audiocoders using our own $V_{mem}$-based backpropagation method\footnote{Table 1: Audio-to-Image A: SNN: $T_{h}=15$, ANN : no quantization for hidden state}\footnote{Table 1: Audio-to-Image B: SNN: $T_{h}=10$, ANN : no quantization for hidden state}. The proposed algorithm brings SNN performance at par with ANNs for the given tasks, thus depicting the effectiveness of the training algorithm. We demonstrate that spiking autoencoders can be used to generate reduced-duration spike maps (``hidden state'') of an input spike train, which are a highly compressed version of the input, and they can be utilized across applications. This is also the first work to demonstrate audio to image synthesis in spiking domain. While training these autoencoders, we made a few important and interesting observations; the first one is the importance of bit masking of the output layer. Trying to steer the membrane potentials of all the neurons is extremely hard to optimize, and selectively correcting only incorrectly spiked neurons makes training easier. This could be applicable to any spiking neural network with a large output layer. Second, while the AE-SNN is trained with spike durations of 15 time steps, we can use hidden state representations of much lower duration to train our audiocoder with negligible loss in reconstruction of images for the audio-to-image synthesis task. In this task, the ANN model trained with Adam outperformed the SNN one when trained with full precision ``hidden state''. However, at ultra-low precision, the hidden state loses it's meaning in ANN domain, but in SNN domain, the network can still learn from it. This observation raises important questions on the ability of SNNs to possibly compute with less data. While sparsity during inference has always been an important aspect of SNNs, this work suggests that sparsity during training can also be potentially exploited by SNNs. We explored how SNNs can be used to compress information into compact spatio-temporal representations and then reconstruct that information back from it. Another interesting observation is that we can potentially train autoencoders and stack them to create deeper spiking networks with greater functionalities. This could be an alternative approach to training deep spiking networks. Thus, this work sheds light on the interesting behavior of spiking neural networks, their ability to generate compact spatio-temporal representations of data, and offers a new training paradigm for learning meaningful representations of complex data. 

\section*{Conflict of Interest Statement}

The authors declare that the research was conducted in the absence of any commercial or financial relationships that could be construed as a potential conflict of interest.

\section*{Author Contributions}
DR, PP, and KR conceived the idea. DR formulated the problem and performed the simulations. DR, PP, and KR analyzed the results. DR wrote the paper.

%\section*{Funding}

\section*{Acknowledgments}
This work was supported in part by the Center for Brain Inspired Computing (C-BRIC), one of the six centers in JUMP, a Semiconductor Research Corporation (SRC) program sponsored by DARPA, the National Science Foundation, Intel Corporation, the DoD Vannevar Bush Fellowship, and by the U.S. Army Research Laboratory and the U.K. Ministry of Defense under Agreement Number W911NF-16-3-0001.

% \section*{Supplemental Data}
%  \href{http://home.frontiersin.org/about/author-guidelines#SupplementaryMaterial}{Supplementary Material} should be uploaded separately on submission, if there are Supplementary Figures, please include the caption in the same file as the figure. LaTeX Supplementary Material templates can be found in the Frontiers LaTeX folder.

\section*{Data Availability Statement}
The datasets analyzed for this study can be found at the following links:
\begin{itemize}
\item MNIST http://yann.lecun.com/exdb/mnist/
\item Fashion MNIST https://github.com/zalandoresearch/fashion-mnist
\item TI-46 audio dataset https://catalog.ldc.upenn.edu/LDC93S9
\end{itemize}
% Please see the availability of data guidelines for more information, at https://www.frontiersin.org/about/author-guidelines#AvailabilityofData

\bibliographystyle{frontiersinSCNS_ENG_HUMS} % for Science, Engineering and Humanities and Social Sciences articles, for Humanities and Social Sciences articles please include page numbers in the in-text citations
\bibliography{references}

%%% Make sure to upload the bib file along with the tex file and PDF
%%% Please see the test.bib file for some examples of references

\end{document}